\documentclass{article}
\usepackage{microtype}
\usepackage{graphicx}
\usepackage{subfigure}
\usepackage{booktabs} % for professional tables
\usepackage{amssymb,amsfonts,amsmath,latexsym,dsfont}
\usepackage{tikz,pgfplots,pgflibraryplotmarks}
\usetikzlibrary{calc}
\usepackage{graphicx}
\usepackage{mathrsfs}
\usepackage{tcolorbox}
\usepackage{pgf,pgfarrows} %% pdf-drawing package

\usepackage{hyperref}

\usepackage[accepted]{icml2019}

\newcommand{\bfC}{{\bf C}}

\newcommand{\bfF}{{\bf F}}

\newcommand{\bfI}{{\bf I}}

\newcommand{\bfK}{{\bf K}}

\newcommand{\bfy}{ {\bf y}}

\newcommand{\bftheta}{{\boldsymbol \theta}}

\newcommand\blfootnote[1]{%
  \begingroup
  \renewcommand\thefootnote{}\footnote{#1}%
  \addtocounter{footnote}{-1}%
  \endgroup
}

\icmltitlerunning{Lean Residual Networks}

\begin{document}

\twocolumn[
\icmltitle{LeanResNet: A Low-cost Yet Effective Convolutional Residual Networks}

\icmlsetsymbol{equal}{*}

\begin{icmlauthorlist}
\icmlauthor{Jonathan Ephrath}{bgu}
\icmlauthor{Lars Ruthotto}{em}
\icmlauthor{Eldad Haber}{eos,xtract}
\icmlauthor{Eran Treister}{bgu}
\end{icmlauthorlist}

\icmlaffiliation{eos}{Department of Earth, Ocean and Atmospheric Sciences, University of British Columbia, Vancouver, Canada}
\icmlaffiliation{xtract}{Xtract AI, Vancouver, Canada}
\icmlaffiliation{em}{Departments of Mathematics and Computer Science, Emory University, Atlanta, GA, USA}
\icmlaffiliation{bgu}{Department of Computer Science, Ben Gurion University, Be'er Sheva, Israel}

\icmlcorrespondingauthor{Eran Treister}{erant@bgu.ac.il}

\icmlkeywords{Mobile networks, reduced parameterized convolutions, residual networks, classification}

\vskip 0.3in
]

%\printAffiliationsAndNotice{} % otherwise use the standard text.

\begin{abstract}
\blfootnote{$^1$Department of Computer Science, Ben Gurion University, Be'er Sheva, Israel. $^2$Departments of Mathematics and Computer Science, Emory University, Atlanta, GA, USA. $^3$Department of Earth, Ocean and Atmospheric Sciences, University of British Columbia, Vancouver, Canada. $^4$Xtract AI, Vancouver, Canada. Corresponding author: Eran Treister: {\tt erant@cs.bgu.ac.il.}}\hspace{-3pt}
Convolutional Neural Networks (CNNs) filter the input data using spatial convolution operators with compact stencils. Commonly, the convolution operators couple features from all channels, which leads to immense computational cost in the training of and prediction with CNNs. 
To improve the efficiency of CNNs, we introduce lean convolution operators that reduce the number of parameters and computational complexity, and can be used in a wide range of existing CNNs.
Here, we exemplify their use in residual networks (ResNets), which have been very reliable for a few years now and analyzed intensively. 
In our experiments on three image classification problems, the proposed LeanResNet yields results that are comparable to other recently proposed reduced architectures using similar number of parameters.
\end{abstract}

\section{Introduction}

Convolution Neural Networks (CNNs)~\cite{LeCun1990} are among the most effective machine learning approaches for processing high-dimensional data and are indispensable in,  e.g., in recognition tasks involving speech~\cite{RainaEtAl2009} and image \cite{KrizhevskySutskeverHinton2012} data. 

In a CNN, the features are grouped into channels.
Through the convolution operators, each feature interacts with other features from a small neighborhood in the same channel and, in most existing approaches, the features from the same neighborhood in the remaining channels; \cite{Gu:2018id,Goodfellow:2016wc}.
A drawback of this fully coupled approach is that the number of convolution operators in a layer is proportional to the product of the number of input and output channels.
This scaling can be expensive when using wide architectures and leads to a large number of weights, often in the millions and beyond. It also complicates the deployment of such CNNs, especially on devices with limited memory resources. 

In recent years there has been an effort to reduce the number of parameters in CNNs. 
Among the different approaches are the
methods of pruning \cite{pruning92,SongHan2015,PrunningCornel2017,Luo_2017_ICCV} and sparsity \cite{wen2016learning,SparsConvCornell2017,SparseReguSongHan2016} that have been typically applied to reduce weights in full networks. It has been shown that once a network is trained, a large portion of its weights can be removed without hampering its efficiency by much. However, the non-zero structure of the weights in the resulting networks is typically unstructured, which may lead to inefficient deployment of the networks on hardware. Still, the success of pruning suggests that there is a significant redundancy in standard CNNs \cite{molchanov2016pruning}.

Another recent effort to reduce the number of parameters in networks is to define architectures based on ``depth-wise'' separable convolutions, which are block diagonal convolution operators.
The depth-wise convolution restricts the interaction of each feature to its nearby features in the same channel. 
To facilitate coupling across the channels, the depth-wise operators is typically used in conjunction with point-wise $1\times1$ convolutions. This was applied in the works of \cite{howard2017mobilenets,sandler2018mobilenetv2,BottleNeckMinWang,SuffleNet,ma2018shufflenet}, together with either bottleneck or shuffling techniques. Since these works use the depth-wise and $1\times 1$ separately, with activation and batch normalization layers in between them, they require a redesign of existing CNN architectures. In addition, applying the depth-wise convolution on its own has a high ratio of floating point operations (FLOPs) to memory access, which has led to the design of networks with shifts instead of convolutions \cite{wu2018shift}. It is known, however, that memory access is the true bottleneck in modern parallel hardware, and not necessarily FLOPs. In fact, the work \cite{qin2018diagonalwise} suggests a superior implementation of the depthwise convolution on GPUs, which is involved with more FLOPs than necessary.

In this paper, we propose a novel way to parameterize CNNs more efficiently, while simply keeping the same structure of the known networks, e.g. residual networks (ResNets) \cite{he2016deep,he2016identity}, which have been one of the most reliable architectures in the literature. Our goal is to reduce the number of weights in the networks and the costs of training and evaluating the CNN. Similarly to recent approaches we use depth-wise convolutions, and $1\times 1$ convolutions to impose coupling between channels. The following three aspects set our work apart from other approaches: (1) We \emph{linearly} add the depth-wise and $1\times 1$ convolutions so that the two operations can be applied simultaneously in hardware, in the same memory read.
    (2) The combined convolution can be simply used as a single convolution operator instead of the standard convolution in any existing CNN, without any structural changes to the architecture.
   (3) We use a 4-point stencil only instead of the standard $3\times3$ or larger stencil, to further reduce memory access and FLOPs of the depth-wise convolution.

\section{ResNets with Lean Convolution Operators} % (fold)
\label{sec:math}

We consider a standard residual network (ResNet) \cite{he2016deep,he2016identity} as a baseline architecture, since it has been very successful and reliable for many tasks. Given a data sample $\bfy_0$, the forward propagation through the network is defined by a series of steps, where the $j$th step is given by
\begin{equation}\label{eq:Resnet}
    \bfy_{j+1} = \bfy_{j} + \bfF(\bftheta_{j},\bfy_j), \quad \text{for} \quad j=0,\ldots,N-1.
\end{equation}
Here, $\bftheta_{j}$ is the set of weights associated with the $j$th step. The nonlinear term in \eqref{eq:Resnet} usually reads
\begin{equation}\label{eq:F}
\bfF(\bftheta,\bfy) = \bfK_2(\bftheta^{(2)}) \sigma( {\cal N}(\bfK_1(\bftheta^{(1)}) \sigma({\cal N}(\bfy)))),
\end{equation}
where $\sigma(x) = \max\{x,0\}$ denotes a element-wise rectified linear unit (ReLU) activation function, the weights are divided into $\bftheta^{(1)}$ and $\bftheta^{(2)}$ that parameterize the two linear operators $\bfK(\bftheta^{(1)})$ and $\bfK(\bftheta^{(2)})$. $\mathcal{N}$ denotes a normalization layer that has trainable parameters as well (omitted here for ease of presentation). The operators $\bfK_1$ and $\bfK_2$ are composed of spatial convolution operators. If the input $\bfy$ has $c_{\rm in}$ channels, and the output $\bfK_1\bfy$ has $c_{\rm out}$ channels, then a common choice for $\bfK_1$ is a $c_{\rm out} \times c_{\rm in}$ block matrix of convolutions, introducing full coupling across the channels.

Our lean convolution operator contains two types of operators. One is the depth-wise (block diagonal) operator which operates on each channel separately, and the other is a $1\times1$ convolution. For example, if $c_{\rm in}=c_{\rm out}=4$, then in matrix form, the operator is given by
\begin{equation}\label{eq:Klean}
	\bfK_{\rm lean} = 	\left(
		\begin{array}{cccc}
			\hat\bfC_1 & \alpha_{1,2}\bfI   &\alpha_{1,3}\bfI &\alpha_{1,4}\bfI \\
			 \alpha_{2,1}\bfI & \hat\bfC_2 &\alpha_{2,3}\bfI &\alpha_{2,4}\bfI   \\
			 \alpha_{3,1}\bfI & \alpha_{3,2}\bfI  & \hat\bfC_3& \alpha_{3,4}\bfI\\
			 \alpha_{4,1}\bfI & \alpha_{4,2}\bfI  &\alpha_{4,3}\bfI & \hat\bfC_4\\	
		 \end{array}
	\right),
\end{equation}
where $\alpha_{i,j}\bfI$ is a scaled identity defined by a learned scalar parameter $\alpha_{i,j}$. The operator $\hat\bfC_i$ is a matrix that corresponds to a 5-point convolution kernel
\begin{equation}\label{eq:5point}
 \left[\begin{matrix}
0 & c_{i,1} & 0 \\ c_{i,2} & \alpha_{i,i} &   c_{i,3} \\
0&  c_{i,4} &0 
\end{matrix}\right],
\end{equation}
where $\alpha_{i,i}$ is the $i,i$ entry of the $1\times1$ convolution, and $c_{i,1},...,c_{i,1}$ are additional 4 parameters per input channel $i$. $\bfK_{\rm lean}$ has $c_{\rm in}\times c_{\rm out} + 4c_{in}$ parameters can be used instead of the standard operators in CNNs.
We note that if the number of input channels is larger than 4, then the $1\times1$ convolution is the dominating operator both in terms of parameters and FLOPs. 

\paragraph{Interpretation} 
 ResNets have been recently interpreted as time-dependent nonlinear PDEs \cite{haber2017stable,Chang2017Reversible,E2017,ChaudhariEtAl2017,lu2018beyond,ruthotto2018deep,chen2018neural}, which  allows the community to analyze and extend ResNets using theoretical and practical ideas from the world of ODEs and PDEs. In this point of view, the depth-wise $3\times3$ convolution  can be seen as a linear combination of a mass term, and discretization of first and second spatial derivatives in each dimension. The $1\times1$ convolution approximates a mass term only. It is known that most simple spatial derivatives can be approximated by a five-point stencil as in \eqref{eq:5point}, and therefore, a $3\times3$ stencil may be unnecessary for extracting features in CNNs. A tremendous advantage will be made in 3D CNNs where the standard 27-point convolutions are replaced with a 7-point stencil (the 3D version of \eqref{eq:5point}).  

\section{Experiments} % (fold)
\label{sec:numerical_examples}
\begin{table*}
    \centering
    \small
	\begin{tabular}{|l|ccc|ccc|ccc|}
\hline
                  & \multicolumn{3}{c|}{CIFAR10} & \multicolumn{3}{c|}{CIFAR100}& \multicolumn{3}{c|}{STL10}\\
Architecture      & Network & Params  & Val. acc.  & Network &Params & Val. acc.  & Network & Params  & Val. acc. \\
\hline
 ResNet          &  A  & 4.3M  &  94.7\%  & C & 27M &  78.5\% &E & 17M  &  84.1\%\\
 ResNet (small)	 &  B    & 0.6M &   91.6\%  &D & 3.8M  &  72.3\%  &F& 1.8M & 78.6\% \\
 MobileNetV2     &A      & 0.5M &  91.5\%  & C   & 2.7M & 71.6\%&  E &1.9M & 80.4\%\\
 ShuffleNetV2    & 0.5x     &0.4M&  89.3\%   & 1.5x  & 2.5M &70.6\%& 1.5x  &2.5M& \textbf{86.2\%}\\
 ShiftResNet        &A      & 0.5M & 92.5\%  &C& 3.1M &  74.2\%& E  & 1.9M & 82.3\%\\
 LeanResNet [ours]      &   A   & 0.5M & \textbf{92.8\%}   &C& 2.9M & \textbf{74.3\%} & E& 2.0M& 83.7\% \\
\hline
	\end{tabular}
		\caption{Classification results }
	\label{tab:Classification}
\end{table*}

We experimentally compare the architectures proposed in this paper to a ResNet with fully-coupled convolutions, and other reduced architectures: ShuffleNetV2 \cite{ma2018shufflenet}, MobileNetV2 \cite{sandler2018mobilenetv2}, and ShiftResNet \cite{wu2018shift}. We use the CIFAR-10, CIFAR100~\cite{krizhevsky2009learning} and STL-10~\cite{coates2011analysis} data sets.
Our primary focus is to compare how the different architectures perform using a relatively small number of weights.
Our experiments are performed with the PyTorch software \cite{paszke2017automatic}.

We adopt a rather standard ResNet architecture, and demonstrate the performance of its lean version. Our ResNet networks consist of several blocks, that are preceded by an opening convolutional $3\times3$ layer, that initially increases the number of channels. Then, there are several blocks, each consisting of a ResNet based part with a number of steps that varies between the different experiments. Each convolution is applied in addition to a ReLU activation and batch normalization as described in \eqref{eq:Resnet}. The last block consists of a pooling layer that averages each channel's map to a single pixel, and we use a fully-connected linear classifier with softmax and cross entropy loss. In Table \ref{tab:config} we summarize the network parameters that we use, which differ in the number of channels and the number of repetitions for each layer.

\begin{table}[h!]
\centering
\small
\begin{tabular}{|c|c|c|}
\hline
Type & Layer width & $\#$ Steps\\
\hline
A & 32-64-128-256 & 2-3-3-3\\
B & 12-24-48-96 & 2-3-3-3\\
C & 64-128-256-512 & 3-5-7-4\\
D & 24-48-96-192 & 3-5-7-4 \\
E & 32-64-128-256-512 & 2-3-3-3-3\\
F & 12-24-48-96-192 & 2-3-3-3-3\\
\hline
\end{tabular}
\caption{Network configurations}
\label{tab:config}
\end{table}

As noted, although the architectures of LeanResNet and ResNet appear to be the same, LeanResNet is based on the parameterized convolution \eqref{eq:Klean}, hence it consumes less parameters.   
The convolution sizes of MobileNetV2 and ShiftResNet were chosen such that the size of the expanded (by 6) $1\times 1$ convolution in a layer is equivalent to the size of a square $1\times1$ convolution of LeanResNet.
The architecture of ShuffleNetV2 is evaluated with the configurations (0.5x,1.0x,1.5x) that were introduced in the papers. 
For training the networks we use the ADAM optimizer \cite{kingma2014adam} and a minibatch of 100. We run 300 epochs and reduce the learning rate by a factor of 0.5 every 75 epochs, starting from 0.1. We also used standard data augmentation, i.e., random resizing, cropping and horizontal flipping.

Our classification results are given in Table \ref{tab:Classification}. The results show that our architecture is in par and in some cases better than other networks. There is no preferred architecture between all options, but our architecture has the advantage of simplicity and resemblance to a standard and reliable ResNet network. We note that although not shown here for a fair comparison, it is better to use $3\times3$ convolutions in the early layers (where there are low numbers of channels and parameters), and then switch to reduced architectures as the network progresses and the number of channels grow.

\subsection{Computational Performance}
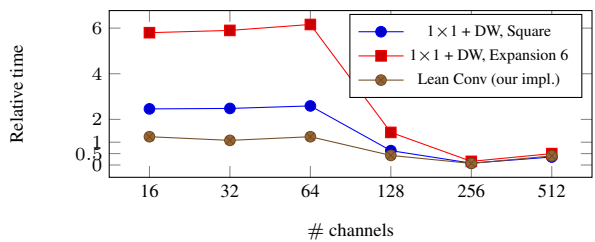
\begin{figure}[h!]
    \centering
    \scriptsize
\begin{tikzpicture}
     \begin{axis}[width=8cm,height=3.8cm,
     xlabel=$\#$ channels,
     ylabel=Relative time,
     %ymode=log,
     xmode=log,
     log ticks with fixed point,
     xtick={16,32,64,128,256,512},
     ytick={0,0.5,1,2,4,6},
     xticklabels from table={labels.dat}{input},
     ]
 \addplot table {
 16   2.46
 32  2.48
 64  2.59
 128  0.63
 256  0.081
 512 0.35
 };
 \addplot table {
 16  5.8 
 32  5.9
 64  6.16
 128 1.43 
 256 0.16 
 512 0.5
 };
 \addplot table {
 16  1.24 
 32  1.08
 64  1.24
 128 0.42
 256 0.07
 512 0.4
 };
\addlegendentry{\tiny{1$\times$1 + DW, Square}}
\addlegendentry{\tiny{1$\times$1 + DW, Expansion 6}}
\addlegendentry{\tiny{Lean Conv (our impl.)}}
 \end{axis}
 \end{tikzpicture}
    \caption{Relative timings of reduced convolutions compared to a $3\times3$ convolution (lower is faster). The expanded and square $1\times1$ convolutions has the same number of parameters. 
    }
    \label{fig:timing}
\end{figure}

We compare the computational cost of our CUDA implementation of the lean convolution with two other combination of layers, comprised of a $1\times1$ convolution that is followed by a depth-wise convolution. In one combination we use $c_{in}=c_{\rm out}$, and in the other $c_{\rm in}\approx 6c_{\rm out}$, but with the same number of parameters. Such layers are applied in \cite{sandler2018mobilenetv2}. We compare the runtime of a typical network: the first layer consists of 16 channels of $512\times 512$ maps, and the maps are coarsened by a factor of 2 when the channels increase by a factor of 2 (i.e., for $512$ channels the images are of size 16). We use a batch size of 64, and compare the runtime of a NVIDIA GeForce 1080Ti GPU for the task. The implementation for the other convolutions is based on PyTorch's $1\times 1$ and grouped convolutions using CUDA 9.2. Figure \ref{fig:timing} summarizes the results. The depthwise convolutions dominate the low channels layers, while all combination converge to the cost of the $1\times1$ convolution as the channels increase (and the depthwise layer becomes negligible). Our implementation of \eqref{eq:Klean} is clearly faster and exploits the simultaneous multiplication of the operators.   
\vspace{-15pt}
\section{Conclusion} \label{sec:discussion}
We present a lean convolution operator that aims at reducing the number of parameters and computational costs of CNNs. In our experiments the new architecture yields classification results that are comparable to other reduced architectures, and is almost as effective as a fully-coupled ResNet. It is important to realize that our new architecture becomes even more advantageous for 3D or 4D problems, e.g., when analyzing time series of medical or geophysical images, the cost of each convolution is much more expensive. Also, the number of weights in the 3D kernels imposes memory-related challenges.

\section{Acknowledgements}
LR’s work is supported by the US National Science Foundation (NSF) awards DMS 1522599 and  DMS 1751636.

\bibliography{LowCostBib}
\bibliographystyle{icml2019}

\end{document}